\documentclass[letterpaper, 10 pt, conference]{ieeeconf}  
\usepackage{graphicx}
\usepackage{caption}
\usepackage{cuted}
\usepackage{amssymb}
\usepackage{bm}  
\usepackage{amsmath}
\usepackage{booktabs}
\usepackage[hidelinks]{hyperref}  

\IEEEoverridecommandlockouts                              

\title{\LARGE \bf
GRASP: Global–Local Representation Alignment and Spatial–Physical Grounding for Bimanual Robotic Manipulation}

\author{
Zeyuan Ren$^{1,2,\dagger}$,
Guanxun Zhao$^{1,3,\dagger}$,
Qi Li$^{1}$,
Hao Wang$^{1}$,
Wenlong Liao$^{1}$,
Keyi Luo$^{1}$, \\
Xijing Zhang$^{1,3}$,
Zhiyuan Zhang$^{1,4}$,
Jingye Cui$^{1}$,
Qibin Gu$^{1,5}$,
Zhichen Wang$^{3}$,
Yucheng Hu$^{1,6}$, \\
Liqiao Xia$^{2}$,
Nan Bu$^{1,7}$,
Zhengde Ma$^{1}$,
Rankun Li$^{1}$,
Xue Yang$^{4}$,
and Wende Xie$^{1,4,*}$
\thanks{
$^{1}$COWARobot Co. Ltd, China.
$^{2}$The Hong Kong Polytechnic University, Hong Kong SAR, China.
$^{3}$Tongji University, Shanghai, China.
$^{4}$Shanghai Jiao Tong University, Shanghai, China.
$^{5}$National University of Singapore, Singapore.
$^{6}$The Chinese University of Hong Kong, Hong Kong SAR, China.
$^{7}$Northeastern University, Shenyang, China.
$^{\dagger}$These authors contributed equally to this work.
$^{*}$Corresponding author: Wende Xie (e-mail: xiewende323@sina.com).
This work was supported by COWARobot Co. Ltd.
}
}
\IEEEaftertitletext{\vspace{-6.0\baselineskip}}
\begin{document}
\maketitle
\thispagestyle{empty}
\pagestyle{empty}
\begin{strip}
    \centering
    \includegraphics[width=\textwidth]{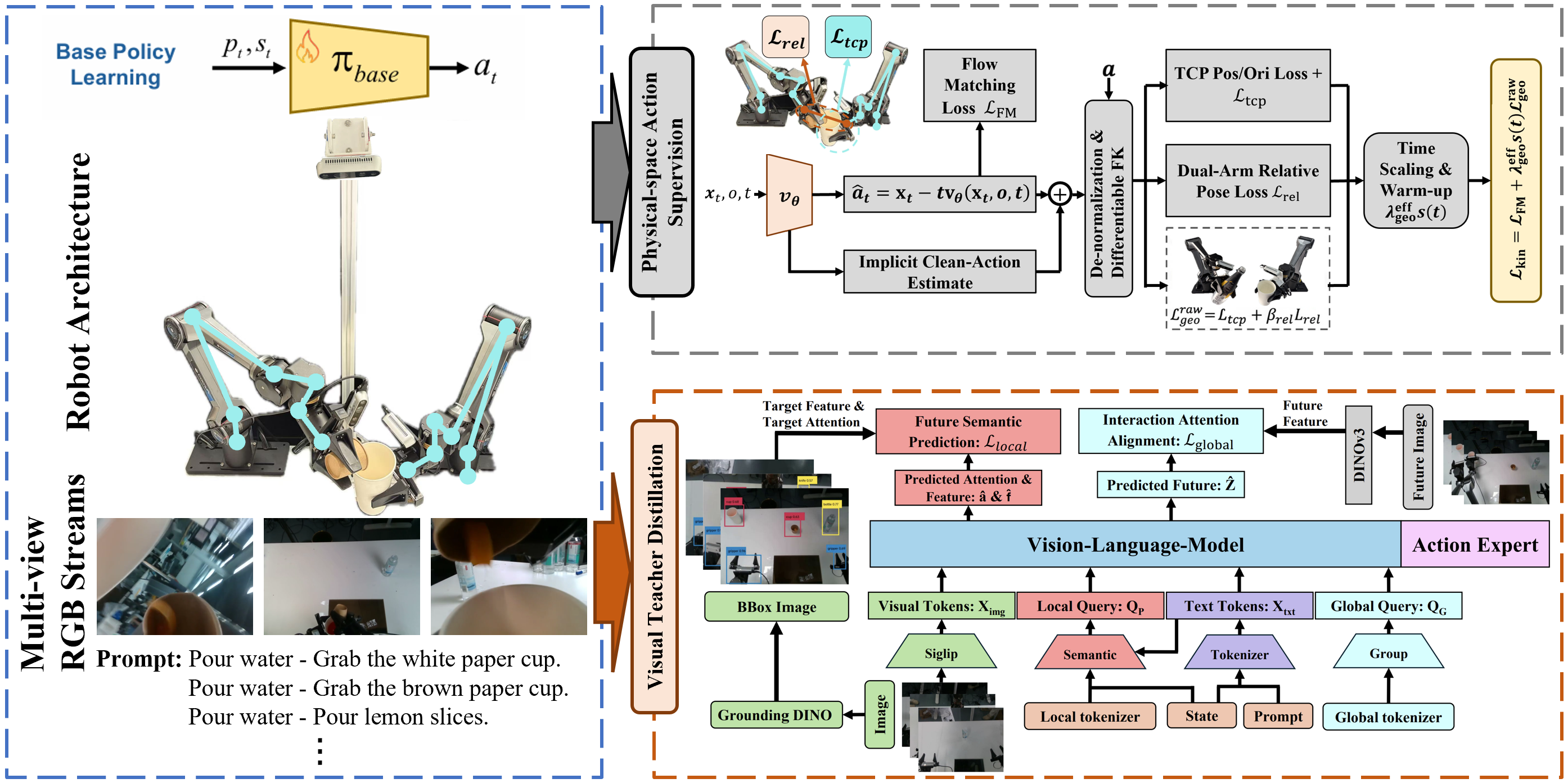}
    \captionof{figure}{Overview of the proposed training-phase prior injection framework. Training-time priors are injected through FK-based physical supervision and DINOv3/Grounding DINO query distillation, all removed at inference.}
    \label{fig:overview}
\end{strip}
\begin{abstract}

Vision-Language-Action (VLA) models that directly regress joint-space actions avoid the deployment-time cost of inverse kinematics. However, their training objectives remain misaligned with task success, as joint-space losses ignore the unequal physical consequences of joint errors and visual conditioning lacks explicit focus on hand-object interaction regions. To address these issues, we propose a training-phase prior injection framework for fine-grained bimanual manipulation. First, a physics-grounded flow-matching loss maps the implicit clean-action estimate at every noise level through differentiable forward kinematics (FK) and supervises the resulting Tool Center Point (TCP) pose and inter-arm relative pose, with a bounded noise-level weighting that compensates gradient decay. Second, we introduce a global prediction query group distilled from a frozen DINOv3 teacher through future-frame feature  prediction, together with a local interaction query supervised by Grounding DINO at the attention level. Both teachers are removed at inference, and the deployment pipeline is that of a standard joint-space flow policy, adding only the learned query pathway. Simulation and real-robot experiments on long-horizon bimanual tasks show that the two priors are complementary across different deployment settings. The full model excels in in-distribution deployment, while the distilled query branch is preferable under strong visual randomization.

\end{abstract}

\section{INTRODUCTION}

Vision-Language-Action models have made significant progress in general robot manipulation, with models such as $\pi_{0.5}$, OpenVLA, and LingBot-VLA 2.0~\cite{pi05,openvla,lingbotvla2} demonstrating strong manipulation capabilities. Flow-based policies such as $\pi_{0}$ generate continuous action sequences through conditional flow matching~\cite{pi0,flowmatching}. Such models either predict the End-Effector Pose (EEP) in task space, requiring a real-time Inverse Kinematics (IK) solver that may yield unreachable poses and discontinuous trajectories due to the one-to-many mapping~\cite{kstar}, or directly regress the joint configuration, which avoids these issues but may misalign the normalized joint-space loss with physical-space task success. Specifically, standard joint-space regression losses do not explicitly account for the configuration-dependent effects of joint errors on end-effector motion. As a result, similar errors in joint space can lead to substantially different task-space deviations after forward-kinematic propagation. Meanwhile, visual conditioning in VLA models is often learned indirectly from the action objective, which may limit consistent attention to fine-grained hand-object interaction regions.

To address these issues, this paper proposes a training-phase prior injection framework for fine-grained bimanual manipulation, along two dimensions: action supervision and visual conditioning.

First, rather than applying kinematic regularization to final actions, we supervise the implicit clean-action estimate throughout the flow trajectory: the estimated clean actions at each noise level are mapped through differentiable forward kinematics, and a geometric loss over TCP pose and inter-arm relative pose ties each penalty to its physical consequence. This loss is training-only, and the policy still outputs joint commands directly at inference.

Second, we append a global query group $Q_{G}$ and a local interaction query $Q_{P}$ to the input sequence, supervised respectively by predicting the dense future-frame features of a frozen DINOv3 teacher~\cite{dinov3} and by attention-level KL divergence distillation using interaction regions derived from Grounding DINO~\cite{groundingdino}, explicitly guiding the model toward hand-object interaction regions. Both teachers are removed at inference.

In our evaluation, we conduct both simulation and real-robot experiments, and show that the proposed framework exhibits a number of desirable properties for fine-grained bimanual manipulation:
\begin{itemize}
\item \textbf{Simulation benchmark:} on 50 bimanual tasks from the RoboTwin 2.0~\cite{robotwin2} benchmark, our full model improves the average success rate over the $\pi_{0.5}$-SFT baseline by 15.1 percentage points in clean settings, while the $\pi_{0.5}$-Global+Local Query variant improves it by 29.0 percentage points under domain randomization.
\item \textbf{Real-robot evaluation:} on four long-horizon real-world tasks, including make watermelon juice, soda bottle opening, beverage mixing, and clothes folding, our method improves the average success rate by 14.2 percentage points over the baseline.
\item \textbf{Physically accurate execution:} supervision in physical space reduces TCP position error compared with joint-space-only training, confirming that the geometric loss better aligns the optimization objective with physical execution.
\item \textbf{Interaction-aware perception:} the decoupled global--local queries improve manipulation performance under randomized scenes and unseen-language instructions by guiding the policy toward task-relevant interaction regions.
\end{itemize}

In summary, our core idea is to compress task-space physical constraints and visual-teacher spatial priors into the training phase. Experiments in both simulation and real-world settings on long-horizon bimanual tasks show consistent improvements in accuracy and robustness.

\section{RELATED WORK}
\subsection{Kinematic constraints in the control of dual-arm robots}

Two common action representations for bimanual manipulation policies are task-space end-effector poses and joint-space configurations. One paradigm predicts the end-effector pose and relies on a cascaded IK solver, but numerical solvers tend to produce discontinuous trajectories near singular configurations. Generative IK methods alleviate this issue through learning~\cite{ikflow}, yet they keep kinematic constraints outside the policy. The other paradigm directly regresses the joint configuration, but dimension-wise regression losses ignore the non-uniformity of the physical consequences of errors propagated through the kinematic chain. Ma \emph{et al.}~\cite{hdp} introduced joint supervision over joint-space and task-space representations, while Lv \emph{et al.}~\cite{kstar} incorporated differentiable forward kinematics to regularize pose learning. Different from these formulations, this paper embeds differentiable forward kinematics directly into the training objective of a joint-space policy, unifying TCP pose and inter-arm relative pose into a geometric loss, requiring no inverse-kinematics solver at inference.

\subsection{Visual Query and Distillation Supervision in VLA models}

Many existing vision-language-action models encode multi-camera observations into dense visual tokens, while the allocation of visual attention is largely learned implicitly from the downstream action objective. Among existing approaches, two directions are particularly relevant. One line introduces learnable query tokens distilled from frozen visual teachers, for example, predicting future scene representations or kinematic evolution from current observations~\cite{vpp,gr2,geopredict}. Such supervision remains at the feature level, without explicitly constraining the spatial allocation of attention. The other line injects spatial priors into attention via open-vocabulary detectors~\cite{owlvit}, but does not distinguish the responsibilities of global scene semantics from local interaction details. In contrast, this paper introduces two dedicated query sets, a global query group and a local interaction query. The former is supervised by predicting the dense features of future frames produced by a self-supervised vision model~\cite{dinov3}, while the latter is supervised through attention-level distillation from attention distributions derived from detection boxes~\cite{groundingdino}. This explicitly guides the model to focus on the hand-object interaction region, with the detector used only during training.

\section{METHODOLOGY}

An overview of the proposed framework is shown in Figure~\ref{fig:overview}. Building on the pretrained $\pi_{0.5}$ backbone, our method integrates two complementary priors during training: a physics-grounded flow-matching loss supervises the implicit clean-action estimate through differentiable forward kinematics, while global--local queries learn future semantics and interaction-region awareness from DINOv3 and offline Grounding DINO supervision. We jointly optimize
\begin{equation}
\mathcal{L}_{\mathrm{total}}=\mathcal{L}_{\mathrm{kin}}
+\lambda_D\mathcal{L}_{\mathrm{global}}
+\lambda_P\mathcal{L}_{\mathrm{local}},
\label{eq:total}
\end{equation}
where $\mathcal{L}_{\mathrm{kin}}$ combines the flow-matching loss with geometric supervision as detailed in Section~\ref{sec:kinematic_supervision}. The global loss $\mathcal{L}_{\mathrm{global}}$ and the local loss $\mathcal{L}_{\mathrm{local}}$ are detailed in Section~\ref{sec:global_local_distillation}. The coefficients $\lambda_D$ and $\lambda_P$ control the contributions of the global and local losses, respectively.

Detector masks provide only training targets, not policy inputs. At inference, image/text encoding, global-query grouping, and task-state conditioning are retained to construct the prefix from current observations. The global and local auxiliary Perceivers remain active to extract the corresponding representations, whereas the DINO teacher, offline region targets, future-feature supervision, and FK loss branch are omitted. The native flow-matching head integrates the predicted vector field from Gaussian noise to generate dual-arm joint actions, without online detection or kinematic post-processing.

\subsection{Differentiable Kinematic Flow Matching Supervision for Bimanual Physical Interaction}
\label{sec:kinematic_supervision}

\begin{figure}[t]
    \raggedright
    \includegraphics[width=\columnwidth]{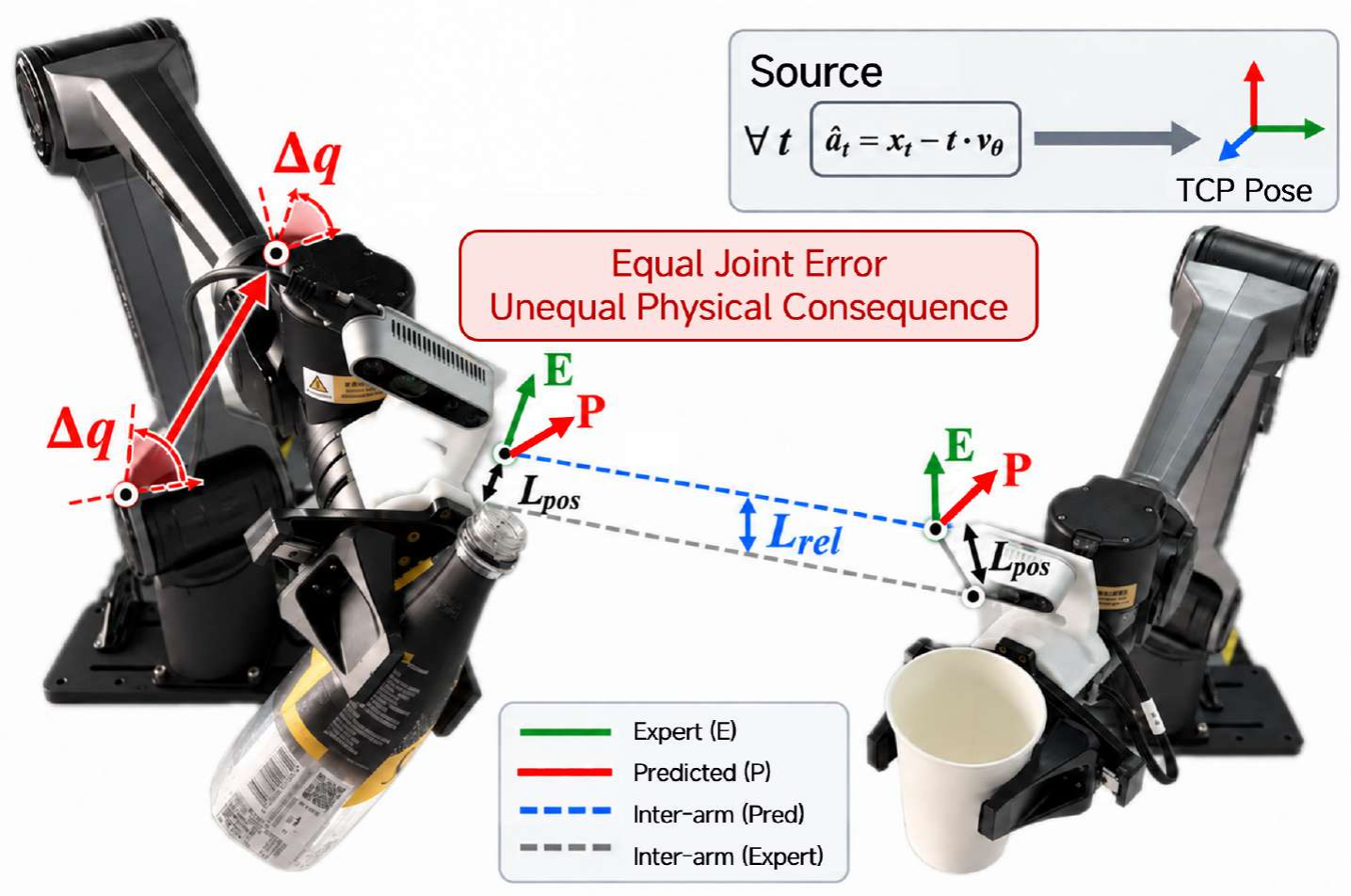}
    \caption{The proposed loss projects intermediate clean actions into task space via differentiable FK and constrains TCP pose and inter-arm relative pose errors, providing physically supervision beyond joint-space regression.}
    \label{fig:GEO}
\end{figure}

Standard joint-space action losses usually treat each joint dimension as an independent regression variable of equal importance. However, for a dual-arm manipulator with serial kinematic chains, different joint errors produce markedly different end-effector effects after being propagated through forward kinematics: shoulder and elbow joints close to the base typically exhibit a larger position-amplification effect, whereas wrist-joint errors manifest more as orientation changes. Therefore, errors of the same magnitude in the normalized joint space do not necessarily correspond to the same TCP position deviation, orientation deviation, or dual-arm relative-pose deviation. To alleviate this inconsistency between the training space and the actual execution space, we introduce a differentiable dual-arm kinematic constraint into the Flow Matching training objective, so that the policy not only learns the action values themselves, but is also directly supervised by their physical consequences, as illustrated in Figure~\ref{fig:GEO}.

Given an observation $o$ and a normalized action sequence $\mathbf{a} \in \mathbb{R}^{H \times 14}$ of length $H$, Conditional Flow Matching~\cite{flowmatching,ddpm} samples $\bm{\epsilon} \sim \mathcal{N}(\mathbf{0}, \mathbf{I})$ and a time variable $t \in (0,1)$, constructing a linear probability path between the action and Gaussian noise.

The policy network predicts the vector field $\mathbf{v}_{\theta}(\mathbf{x}_{t}, o, t)$. At each Flow noise level, we construct the corresponding clean-action estimate:
\begin{equation}
\mathbf{x}_{t} = (1-t)\,\mathbf{a} + t\,\bm{\epsilon}
\end{equation}
\begin{equation}
\mathbf{u}_{t} = \bm{\epsilon} - \mathbf{a}
\quad \Rightarrow \quad
\widehat{\mathbf{a}}_{t} = \mathbf{x}_{t} - t\,\mathbf{v}_{\theta}(\mathbf{x}_{t}, o, t).
\end{equation}
The base flow matching loss is
\begin{equation}
\mathcal{L}_{FM} = \mathbb{E}_{t,\, \bm{\epsilon},\, \mathbf{a}} \left\| \mathbf{v}_{\theta}(\mathbf{x}_{t}, o, t) - \mathbf{u}_{t} \right\|_{2}^{2}.
\end{equation}

When $\mathbf{v}_{\theta}$ accurately approximates the target vector field, $\widehat{\mathbf{a}}_{t}$ approaches the true action $\mathbf{a}$ and can be regarded as the implicit final action at the current noise level. Similar to MDM~\cite{mdm} and RFdiffusion~\cite{rfdiffusion}, this design imposes geometric supervision in data space, but we retain the velocity-field parameterization and obtain $\widehat{\mathbf{a}}_{t}$ analytically, directly from the linear path.

The training objective of this branch is
\begin{equation}
\mathcal{L}_{\mathrm{kin}} = \mathcal{L}_{FM} + \lambda_{\mathrm{geo}}^{eff}\, w(t)\,  \mathcal{L}_{\mathrm{geo}}^{raw}
\end{equation}
\begin{equation}
\lambda_{\mathrm{geo}}^{eff} = \lambda_{\mathrm{geo}} \min\!\left(1, \frac{k}{K_{warmup}}\right).
\end{equation}
Here $\mathcal{L}_{FM}$ is the base flow matching loss defined above, $\mathcal{L}_{\mathrm{geo}}^{raw}$ is the physical-space geometric loss composed below, $w(t)$ is a noise-level weight introduced later in this section, $\lambda_{\mathrm{geo}}$ is the target geometric weight, and $k$ and $K_{warmup}$ denote the current training step and the warm-up duration in steps, respectively, so that $\lambda_{\mathrm{geo}}^{eff}$ is the effective geometric weight after linear warm-up.

Prior work applies differentiable kinematic constraints to the final outputs or key poses of the policy~\cite{kstar,hdp}, or uses them as a standalone generative prior~\cite{ikflow}. Building on the above differentiable kinematic mapping, we extend such physical-space supervision to the implicit clean-action estimate at every Flow noise level, and compose the physical-interaction geometric loss as
\begin{equation}
\mathcal{L}_{\mathrm{geo}}^{raw} = \mathcal{L}_{\mathrm{TCP}} + \beta_{\mathrm{rel}}\, \mathcal{L}_{\mathrm{rel}}.
\end{equation}

The rotation error is the squared geodesic distance $d_{SO(3)}^{2} = \theta^{2}$ on $SO(3)$, the group of $3{\times}3$ rotation matrices, where $\theta$ is the angle of the relative rotation between prediction and ground truth~\cite{zhou6d}. To avoid the unstable gradients of $\arccos$ near $\theta \approx 0$, $\theta^{2}$ is evaluated through a series expansion in $d = 1 - \cos\theta$.

\begin{align}
\mathcal{L}_{\mathrm{TCP}} &= \frac{1}{H} \sum_{h} \Big[ \big\| \Delta\mathbf{p}_{t,h}^{j} \big\|_{2}^{2} + \rho_{\mathrm{geo}}\, \big( \theta_{t,h}^{j} \big)^{2} \Big],\\
\mathcal{L}_{\mathrm{rel}} &= \frac{1}{H} \sum_{h} \Big[ \big\| \Delta\mathbf{p}_{\mathrm{rel},t,h} \big\|_{2}^{2} + \rho_{\mathrm{rel}}\, \big( \theta_{\mathrm{rel},t,h} \big)^{2} \Big],
\end{align}
where $\Delta\mathbf{p} = \widehat{\mathbf{p}} - \mathbf{p}$ and
$\theta = d_{SO(3)}(\widehat{\mathbf{R}}, \mathbf{R})$ denote the position
and rotation residuals, respectively. Here, $j \in \{L,R\}$ indexes the two arms,
and $\mathcal{L}_{\mathrm{TCP}}$ is evaluated separately for each arm in its
corresponding local base frame; the two arm-wise TCP losses are then averaged
when computing $\mathcal{L}_{\mathrm{geo}}^{raw}$.
Hatted quantities are predicted from the arm-joint dimensions of
$\widehat{\mathbf{a}}_{t}$ via differentiable forward kinematics, while
non-hatted quantities are ground truths computed from $\mathbf{a}$.
The TCP term penalizes absolute end-effector pose errors, suppressing error
amplification along the kinematic chains.
The relative-pose term supervises the relative transform between the two hands: the right-arm TCP is first mapped into the left base frame via the calibrated inter-base transformation and then composed with the inverse of the left TCP pose, i.e., $\mathbf{T}_{\mathrm{rel}} = (\mathbf{T}_{L})^{-1}\,{}^{B_L}\mathbf{T}_{B_R}\,\mathbf{T}_{R}$, so that $\Delta\mathbf{p}_{\mathrm{rel}}$ and $\theta_{\mathrm{rel}}$ directly constrain the distance, direction, and relative rotation between the two hands.
$\rho_{\mathrm{geo}}$ and $\rho_{\mathrm{rel}}$ weight orientation errors
against position errors, and $\beta_{\mathrm{rel}}$ balances the relative-pose
term against the TCP term.

Backpropagated through the differentiable forward kinematics, the gradient of each joint is weighted by the transpose of its Jacobian, so joints with a larger influence on the end-effector pose receive stronger effective supervision, correcting the unequal physical consequences of joint errors noted at the beginning of this section.

Since the derivative of $\widehat{\mathbf{a}}_{t} = \mathbf{x}_{t} - t\,\mathbf{v}_{\theta}$ with respect to the velocity prediction contains the factor $t$, the geometric gradients weaken in the low-$t$ region, yet $\widehat{\mathbf{a}}_{t}$ at the low-noise end is usually more reliable, so the supervision strength is distributed oppositely to the reliability. The geometric computation uses only the 12 arm-joint dimensions of the estimated clean action $\widehat{\mathbf{a}}_{t}$, since the gripper dimensions do not participate in the FK computation, while the Flow Matching loss still supervises the full 14-dimensional action. The clean-action estimation error satisfies
\begin{equation}
\widehat{\mathbf{a}}_{t} - \mathbf{a} = -t \left( \mathbf{v}_{\theta}(\mathbf{x}_{t}, o, t) - \mathbf{u}_{t} \right).
\end{equation}

Under a small prediction error and a first-order local linearization of FK, the squared geometric loss and its gradient with respect to the velocity residual scale approximately as $t^{2}$; physical supervision in the low-$t$ region is therefore suppressed.

Drawing on the noise-level weighting ideas of EDM and recent Min-SNR-style weighting analyses for flow training~\cite{edm,meanflow26}, we introduce a bounded time-scaling function:
\begin{equation}
w(t) = \mathbf{1}\!\left[ t > t_{\min} \right] \min\!\left( w_{\max}, \frac{1}{t^{2} + \delta} \right).
\end{equation}
This function compensates for the gradient decay caused by the factor $t$. The indicator function switches off supervision when $t \leq t_{\min}$, $w_{\max}$ caps the upper bound, and $\delta$ ensures stability. In our implementation, $t_{\min} = 0.05$, $w_{\max} = 10$, and $\delta = 10^{-3}$. We keep the original sampling distribution and only correct the gradient scale through the loss weight.

To avoid overly strong geometric gradients at the beginning of training, we linearly warm up the geometric weight with the training step $k$ up to $K_{warmup}$, thereby gradually strengthening the physical constraint without disrupting the original action distribution.

\subsection{Interaction-Guided Global--Local Query Distillation}
\label{sec:global_local_distillation}
\begin{figure}[t]
    \raggedright
    \includegraphics[width=\columnwidth]{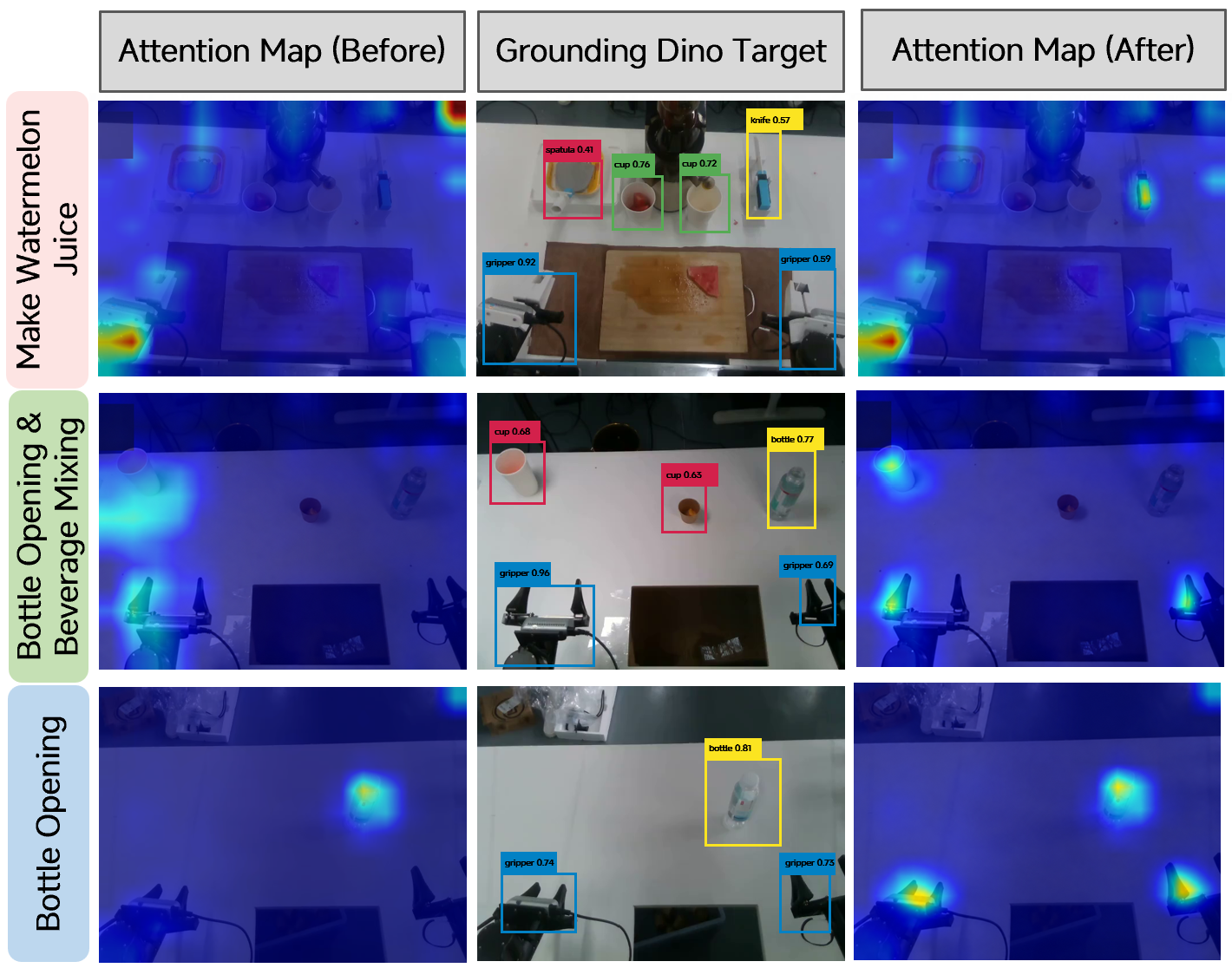}
    \caption{Interaction-region-guided attention distillation. Left to right: attention before distillation, Grounding DINO region targets, and attention after distillation. Local queries focus attention on task-relevant objects and grippers.}
    \label{fig:DINO}
\end{figure}

The native $\pi_{0.5}$ encodes multi-view images, language instructions, and the robot state into a unified condition, and generates joint action sequences through conditional flow matching. Although the pretrained vision-language backbone has strong scene understanding capability, its allocation of visual information is mainly driven indirectly by the action regression objective. In fine-grained manipulation, the grippers and the currently relevant objects occupy only a small portion of the image, while long-horizon tasks may contain multiple distractor objects. We therefore introduce a global--local query design, illustrated in Figure~\ref{fig:DINO}: a global query group $Q_G$ provides future-scene semantic supervision, whereas a local manipulation query $Q_P$ uses offline interaction-region priors to emphasize the objects and grippers relevant to the current action.

At time $\tau$, following the native $\pi_{0.5}$ input formulation, we encode the current multi-view images into visual tokens $X_{\mathrm{img}}$ and the task instruction together with the robot state into text tokens $X_{\mathrm{txt}}$. We augment this prefix with global and local queries while keeping the original action suffix unchanged:
\begin{equation}
X_{\mathrm{pre}}=\mathrm{concat}(X_{\mathrm{img}},X_{\mathrm{txt}},Q_G,Q_P),
\label{eq:prefix}
\end{equation}
where $Q_G\in\mathbb{R}^{K_G\times D}$ consists of $K_G$ global queries obtained by grouping learnable sampler latents, and $Q_P\in\mathbb{R}^{S\times D}$ consists of $S$ local queries obtained by task-state conditioning of learnable local tokens. Here $D$ is the VLM hidden width, so both query sets share the feature dimension of $X_{\mathrm{img}}$ and $X_{\mathrm{txt}}$. This prefix width is distinct from the teacher feature width $D_T$ used for auxiliary prediction. We detail the global and local branches in turn.

For the global branch, we partition a shared pool of $N_R$ learnable sampler latents $R=\{r_i\}_{i=1}^{N_R}$ into $K_G$ equal-sized groups, each containing $n_g=N_R/K_G$ latents, with $N_R$ divisible by $K_G$. We average each group to obtain
\begin{equation}
q_G^k=\frac{1}{n_g}\sum_{i=(k-1)n_g+1}^{kn_g}r_i,
\quad k=1,\ldots,K_G.
\label{eq:group}
\end{equation}
The resulting global prefix tokens $Q_G=\{q_G^k\}_{k=1}^{K_G}$ receive gradients from both the action and future-feature objectives, while the full latent pool $R$ serves as the query set of the auxiliary perceiver sampler~\cite{perceiverio}. 

To encourage the global queries to capture information relevant to future scene evolution, we train this sampler to predict future visual-semantic features from the current observation. The supervision targets are derived from a frozen DINOv3 ViT-L/16 teacher:
\begin{equation}
Z_{\tau+\Delta}=\mathrm{sg}(E_D(I_{\tau+\Delta}))
\in\mathbb{R}^{N_D\times D},
\label{eq:teacher}
\end{equation}
where $I_{\tau+\Delta}$ is the image $\Delta$ frames ahead, $E_D$ is the teacher encoder, $N_D$ and $D$ denote its patch count and feature width, and $\mathrm{sg}$ denotes stop-gradient. The teacher features are resampled to $N_R$ prediction slots to match the sampler output, yielding targets of shape $N_R\times D$.

In our implementation, $N_R=576$ and $K_G=8$. Teacher images are resized and center-cropped to $224\times224$, yielding $N_D=196$ patch features of width $D=1024$, with a default future offset of $\Delta=30$ frames.

After the vision-language backbone processes $X_{\mathrm{pre}}$, we extract the output hidden states at the image-token and global-query positions, denoted by $H_{\mathrm{img}}\in\mathbb{R}^{N\times D}$ and $H_G\in\mathbb{R}^{K_G\times D}$, respectively, where $N$ is the number of image tokens. In particular, $Q_G$ denotes the input global queries, whereas $H_G$ denotes their contextualized representations after interaction with the other prefix tokens. Their concatenation $[H_{\mathrm{img}};H_G]$ along the token dimension provides the key-value context of the auxiliary Perceiver, while the full learnable latent pool $R$ provides its queries:
\begin{equation}
\hat{Z}_{\tau+\Delta}=\mathrm{Perceiver}\big(Q=R,\;KV=[H_{\mathrm{img}};H_G]\big).
\label{eq:perceiver}
\end{equation}
The future semantic objective is a masked per-sample regression loss:
\begin{equation}
\mathcal{L}_{\mathrm{global}}=M_\tau\left\|\hat{Z}_{\tau+\Delta}-Z_{\tau+\Delta}\right\|_F^2,
\label{eq:dino}
\end{equation}
where $M_\tau$ suppresses invalid future frames. This branch predicts a future visual representation from the current observation; it does not use the future image at inference time.

For the local branch, we construct $S$ task-state-conditioned local prefix tokens $Q_P\in\mathbb{R}^{S\times D}$ from learnable query slots and the current instruction and robot state.

For each slot, we add instruction and robot-state features to a learnable vector $p_j\in\mathbb{R}^{D}$:
\begin{equation}
U_j^{(0)}=p_j+W_t\operatorname{Task}(X_{\mathrm{txt}})+W_s s_\tau,
\quad j=1,\ldots,S,
\label{eq:local_init}
\end{equation}
where $\operatorname{Task}(X_{\mathrm{txt}})$ extracts the task-instruction representation from $X_{\mathrm{txt}}$ while excluding robot-state and padding tokens, and $s_\tau$ is the current normalized robot proprioceptive state. The learned projections $W_t$ and $W_s$ map these two inputs to $D$-dimensional vectors. Each slot has its own learnable vector $p_j$, while all slots share the same instruction and state conditioning. The slot count $S$ is fixed within each experiment. The resulting $Q_P=\{U_j^{(0)}\}_{j=1}^{S}$ is concatenated with the other prefix tokens and processed by the $\pi_{0.5}$ backbone.

Thus, $Q_P$ provides task-state-conditioned local query tokens to the VLM backbone. After the VLM backbone, we reuse the Perceiver query-to-context structure for local feature extraction. Specifically, the local tokens serve as queries and the VLM image and global-query hidden states provide the key-value context:
\begin{equation}
Y_P=\operatorname{Perceiver}
\left(Q=Q_PW_Q,\;KV=[H_{\mathrm{img}};H_P]\right),
\label{eq:local_perceiver}
\end{equation}
where $H_P\in\mathbb{R}^{S\times D}$ and $H_{\mathrm{img}}\in\mathbb{R}^{N\times D}$ denote the VLM output hidden states corresponding to the local prefix tokens $Q_P$ and the $N$ image tokens, respectively.

Grounding DINO provides offline spatial supervision for the local queries. For sample $b$ and slot $j$, the target attention $a_{bj}^*$ is constructed by mapping subtask-relevant, confidence-weighted detections to the visual-token grid and normalizing the resulting spatial weights; this target is detached from gradient computation. The predicted attention $\hat a_{bj}$ is obtained from the final cross-attention layer of the local Perceiver in Eq.~\ref{eq:local_perceiver}, averaged over heads and renormalized over valid image-token positions. These distributions describe where the query should attend and where it actually attends, respectively.

For sample $b$, we denote the corresponding image hidden states by $H_{\mathrm{img},b}$. The target and predicted visual features are obtained by attention-weighted pooling:
\begin{equation}
f_{bj}^*=\operatorname{sg}\!\left(\operatorname{Pool}(a_{bj}^*,H_{\mathrm{img},b})\right),
\hat f_{bj}=\operatorname{Pool}(\hat a_{bj},H_{\mathrm{img},b}),
\label{eq:local_targets}
\end{equation}
where $\operatorname{Pool}$ aggregates the same VLM image hidden states using the target or predicted attention, yielding $f_{bj}^*,\hat f_{bj}\in\mathbb{R}^{D}$, and $\operatorname{sg}$ stops gradients through the target feature. Let $g_{bj}\in\{0,1\}$ indicate valid local supervision and $G=\max(1,\sum_{b,j}g_{bj})$. The local objective is
\begin{equation}
\begin{aligned}
\mathcal L_{\mathrm{attn}}&=\frac{1}{G}\sum_{b,j}g_{bj}
\mathrm{KL}(a_{bj}^*\|\hat a_{bj}),\\
\mathcal L_{\mathrm{feat}}&=\frac{1}{GD}\sum_{b,j}g_{bj}
\|\hat f_{bj}-f_{bj}^*\|_2^2,\\
\mathcal L_{\mathrm{local}}&=\mathcal L_{\mathrm{attn}}+\eta\mathcal L_{\mathrm{feat}}.
\end{aligned}
\label{eq:local_loss}
\end{equation}
where the KL term aligns local attention with the interaction-region prior, the feature term preserves the corresponding visual content, and $\eta\geq0$ balances the two terms. The factor $1/G$ averages over valid sample-slot pairs, while the additional factor $1/D$ averages the squared feature error over the $D$ VLM feature dimensions. Invalid targets contribute zero local loss, while the local prefix tokens and their task-state projection layers also receive gradients from the action objective.

\section{EXPERIMENTS}

Our experiments evaluate the effect of the geometric supervision on action accuracy, of the query distillation on robustness under randomized scenes and unseen language, and of their combination against the baseline VLA model.

\subsection{Simulation Setup}

We first evaluate our method on the RoboTwin 2.0 benchmark~\cite{robotwin2,robotwin}, which provides a diverse set of robot manipulation tasks with different object configurations and environmental variations.
Following the RoboTwin 2.0 evaluation protocol, we conduct experiments on 50 manipulation tasks under two evaluation conditions: clean scenes with seen-language instructions and randomized scenes with unseen-language instructions.

The clean setting evaluates the basic manipulation capability under standard environments, while the randomized setting introduces visual and environmental variations, including background changes, distractor objects, lighting variations, and scene perturbations, together with unseen-language instructions, providing a more challenging evaluation scenario.

All simulation experiments use the Aloha AgileX dual-arm embodiment provided by RoboTwin 2.0.
For each task, we fine-tune a separate policy from the pretrained $\pi_{0.5}$ checkpoint on 50 clean demonstrations generated by the RoboTwin 2.0 data pipeline, matching the official benchmark protocol.
The baseline model is trained with RGB observations using a batch size of 256 for 20,000 training steps.

During evaluation, we use a fixed seed--prompt index and count 100 episodes per task and condition; episodes with initialization failures are replaced from the candidate list.
The success rate is determined by the task-specific success criterion provided by RoboTwin 2.0.
We report the average success rate over all evaluated tasks.

\subsection{Baselines and Implementation Details}

To investigate the contribution of each proposed component, we compare five variants:

\begin{itemize}
    \item $\pi_{0.5}$-SFT: the baseline fine-tuned from $\pi_{0.5}$ with RGB observations.
    \item $\pi_{0.5}$-Global Query: the baseline augmented with the global prediction query group only.
    \item $\pi_{0.5}$-Global+Local Query: the baseline augmented with both the global prediction query group and the local interaction query.
    \item $\pi_{0.5}$-Geometry: the baseline incorporating the proposed geometric supervision.
    \item $\pi_{0.5}$-Geometry+Global+Local Query (Ours): the complete model combining geometric supervision and both queries.
\end{itemize}

All variants are initialized from the same pretrained $\pi_{0.5}$ checkpoint and fine-tuned on the same demonstrations for the same number of training steps.
Position errors are measured in meters, and rotation errors are measured in radians as the geodesic angle on $SO(3)$.
The rotation weights follow a characteristic-length conversion $\rho{=}r^{2}$:
$\rho_{\mathrm{geo}}{=}0.3$ corresponds to $r{\approx}0.55$\,m, on the order of the arm's manipulation radius,
and $\rho_{\mathrm{rel}}{=}0.1$ corresponds to $r{\approx}0.32$\,m, the typical inter-hand distance during bimanual coordination,
so that one radian of orientation error is penalized comparably to the induced positional displacement at the corresponding scale.
The geometric branch enters as an auxiliary objective at weight $\lambda_{\mathrm{geo}}{=}0.15$,
with the relative-pose term weighted at $\beta_{\mathrm{rel}}{=}0.1$ of the TCP term.
The future-feature loss is high-dimensional and is scaled down by $\lambda_{D}{=}0.08$ to
keep its gradient contribution below the flow-matching objective, while the attention KL
term is dimensionless and added unweighted with $\lambda_{P}{=}1$.
We linearly warm up the geometric weight over the first $K_{warmup}{=}2{,}000$ steps,
10\% of the total training steps, during which $\mathcal{L}_{FM}$ undergoes its initial rapid descent.
The action horizon is $H{=}50$, corresponding to approximately 1.67\,s at 30\,Hz, which contains the
$\Delta{=}30$-frame future-prediction target within the executed chunk; flow integration
uses 10 steps at inference.
All hyperparameters are selected via a coarse sweep on held-out tasks and fixed for all reported experiments.

\subsection{Simulation Results}

\begin{table*}[t]
\centering
\caption{
Simulation results on the RoboTwin 2.0 benchmark.
Success rates (SR) are averaged over 50 manipulation tasks with 100 episodes per condition.
TCP err. denotes the mean and maximum deviation between the executed end-effector trajectory and the expert trajectory.
$^{\dagger}$Results reported by the RoboTwin 2.0 benchmark~\cite{robotwin2} under its official protocol: Aloha AgileX embodiment, 50 clean demonstrations per task, and 100 rollouts per condition. Action Chunking with Transformers (ACT)~\cite{act} and Diffusion Policy (DP)~\cite{diffusionpolicy} are trained from scratch; RDT~\cite{rdt} and $\pi_{0}$~\cite{pi0} are fine-tuned from released pretrained weights in the single-task setting.
}
\label{tab:robotwin2_results}
\begin{tabular}{lcccc}
\toprule
Method & Clean SR(\%) $\uparrow$ & Randomized SR(\%) $\uparrow$ & TCP err. mean(mm) $\downarrow$ & TCP err. max(mm) $\downarrow$\\
\midrule
ACT$^{\dagger}$ & 29.7 & 1.7 & -- & --\\
DP$^{\dagger}$ & 28.0 & 0.6 & -- & --\\
RDT$^{\dagger}$ & 34.5 & 13.7 & -- & --\\
$\pi_{0}$$^{\dagger}$ & 46.4 & 16.3 & -- & --\\
\midrule
$\pi_{0.5}$-SFT & 64.13 & 36.15 & 5.3 & 7.7\\
$\pi_{0.5}$-Global Query & 64.26 & 55.29 & 5.1 & 7.2\\
$\pi_{0.5}$-Global+Local Query & 68.30 & \textbf{65.15} & 4.7 & 7.1\\
$\pi_{0.5}$-Geometry & 67.74 & 41.24 & \textbf{4.6} & 6.8\\
$\pi_{0.5}$-Geometry+Global+Local Query (Ours) & \textbf{79.22} & 52.84 & \textbf{4.6} & \textbf{5.7}\\
\bottomrule
\end{tabular}
\end{table*}

Table~\ref{tab:robotwin2_results} summarizes the simulation results on RoboTwin 2.0.
For reference, the official benchmark reports average success rates of 34.5\%/13.7\% for Robotics Diffusion Transformer (RDT)~\cite{rdt} and 46.4\%/16.3\% for $\pi_{0}$~\cite{pi0} under its Easy/Hard settings~\cite{robotwin2}, which correspond to our Clean/Randomized conditions, respectively.
Our $\pi_{0.5}$-SFT baseline, built on a more recent backbone, already surpasses these results, and the proposed framework further improves over it.
The baseline $\pi_{0.5}$-SFT achieves an average success rate of 64.13\% on clean tasks and 36.15\% on randomized tasks.

Introducing the proposed geometric supervision improves the clean-task performance from 64.13\% to 67.74\%, demonstrating that physical constraints provide additional guidance for improving action accuracy and task execution.
For example, tasks requiring precise object positioning, such as \textit{Move Pillbottle Pad} and \textit{Pick Dual Bottles}, benefit substantially from geometric supervision. The TCP error columns further show that the geometric supervision reduces the average deviation from the expert end-effector trajectory from 5.3\,mm to 4.6\,mm and the maximum deviation from 7.7\,mm to 6.8\,mm, confirming improved physical execution accuracy.

The global--local query distillation provides larger improvements under the randomized setting.
Adding the global prediction query group alone leaves clean performance almost unchanged (64.13\% $\rightarrow$ 64.26\%) but already improves randomized performance from 36.15\% to 55.29\%, and further adding the local interaction query ($\pi_{0.5}$-Global+Local Query) brings the randomized-task performance to 65.15\% and clean performance to 68.30\%, indicating that the distilled global--local queries help the policy better focus on task-relevant regions under varied scene appearances and unseen-language instructions.

We hypothesize that the pronounced gain under the randomized setting stems from the complementary invariance properties of the two supervision signals.
First, the DINOv3 teacher is trained with strong appearance-level augmentations, and its dense features are largely invariant to texture, lighting, and background changes; predicting such features likely encourages the policy to condition its actions on appearance-invariant scene semantics rather than raw pixel statistics.
Second, the attention-level supervision anchors the local interaction query on the hand--object region, so that distractors and background perturbations outside the interaction region are explicitly down-weighted.
Third, both supervision signals are derived from visual teachers and detection boxes, which do not depend on instruction phrasing; the distilled queries may thus provide a task-relevant visual pathway that remains effective when language embeddings drift out of the training distribution under unseen-language instructions.

By combining both components, our full model achieves the best clean-scene performance, reaching 79.22\% success rate, 15.1 percentage points above the baseline and 10.9 points above the query-only variant, while obtaining 52.84\% on randomized scenes, a 16.7-point improvement over the baseline.
An interaction effect is observed between the two priors. On clean scenes, geometric supervision and query distillation are super-additive: the gain of the full model exceeds the sum of the individual gains by 7.3 points. Under domain randomization, however, the full model trails the query-only variant by 12.3 points; one possible explanation is that the geometric objective steers the policy toward precise reproduction of expert kinematics, which improves in-distribution fitting but reduces reliance on the appearance-invariant visual cues provided by the distilled queries. The two priors are therefore complementary across deployment regimes: the full model is best suited to real-world, approximately in-distribution deployment, while the query-only variant is preferable under strong visual randomization.

\subsection{Real-Robot Experiments}

\subsubsection{Experimental Setup}

We validate our method on a real-world dual-arm PiPER-X robot with parallel grippers (Fig.~\ref{fig:real_setup}).
Three Intel RealSense D435i cameras provide multi-view observations; only RGB images are used for policy inference, which runs at 30\,Hz.
For the make watermelon juice task, 3D-printed grooved gripper attachments improve the grasping stability on the knife and spatula.
For each task, we collect about 150 demonstrations. Each method is trained with 5 random seeds and evaluated over 30 trials per checkpoint; we report the mean success rate across seeds with error bars denoting one standard deviation, where a trial is successful only if the entire task sequence is completed without human intervention.

\subsubsection{Long-Horizon Manipulation Tasks}

We evaluate the proposed method on four representative real-world manipulation tasks, including make watermelon juice, soda bottle opening, beverage mixing, and clothes folding.
These tasks involve long-horizon execution, tool usage, liquid manipulation, and bimanual coordination.

\textbf{Make watermelon juice.}
The robot cuts the watermelon with a knife, transfers the pieces into the juicer with the spatula, and sets the extracted juice aside for subsequent tasks.

\textbf{Soda bottle opening.}
The robot twists off the cap in two stages, discards it, and places the opened bottle on the table.

\textbf{Beverage mixing.}
The robot pours the ingredient and approximately half of the soda into the juice cup, discards the empty containers, and places the mixed beverage at the target position.

\textbf{Clothes folding.}
The robot spreads the clothes flat, folds the lower and upper edges inward, folds along the vertical center axis, and transfers the folded clothes to the target location.

\subsubsection{Real-World Results}
\begin{figure}[t]
    \raggedright
    \includegraphics[width=\columnwidth]{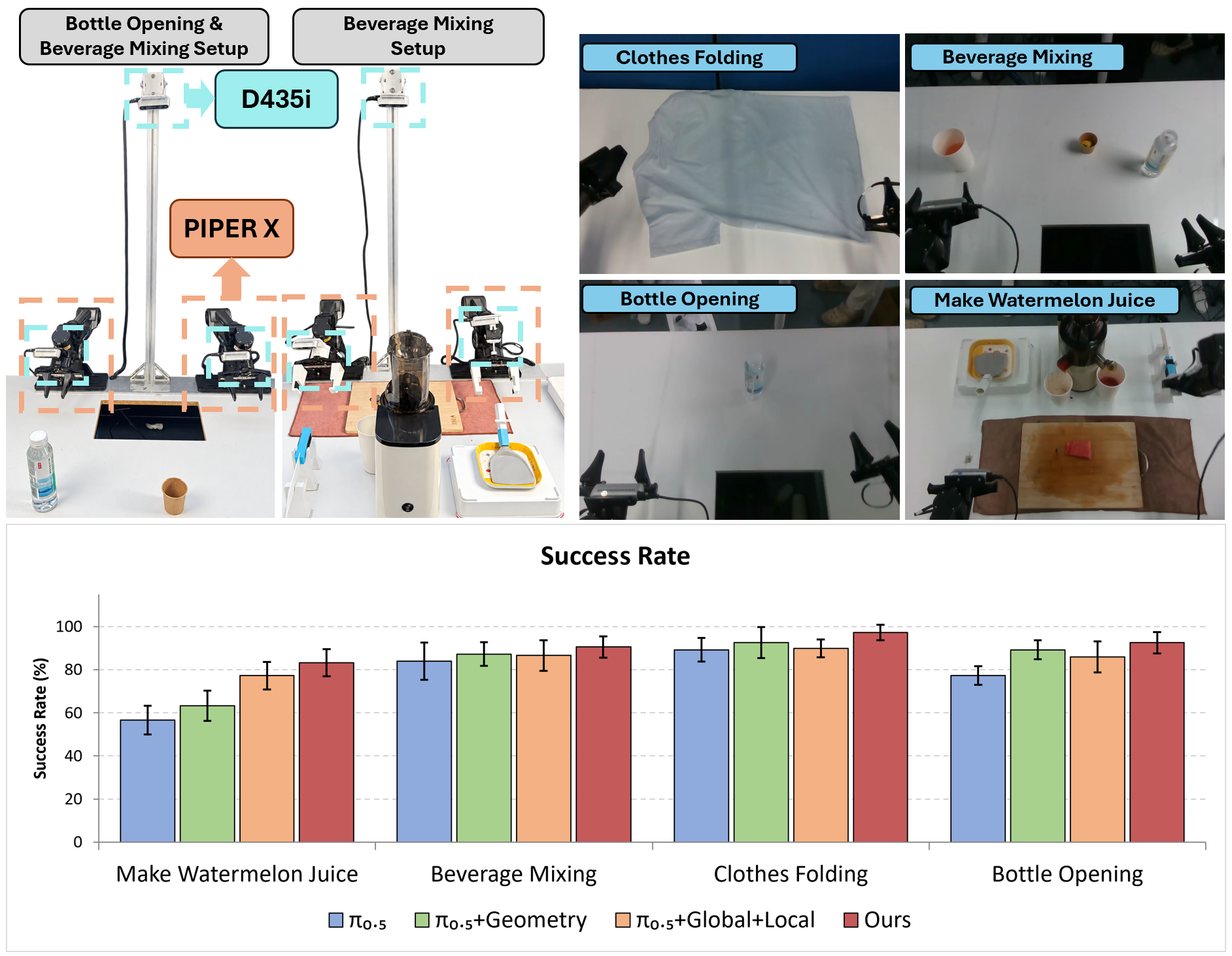}
    \caption{Real-world manipulation setup and results on the dual-arm PiPER-X robot. Each method is trained with 5 random seeds and each checkpoint is evaluated over 30 independent trials; bars show the mean success rate across seeds and error bars denote one standard deviation.}
    \label{fig:real_setup}
\end{figure}
Figure~\ref{fig:real_setup} summarizes the real-world deployment results.

Compared with the $\pi_{0.5}$-SFT baseline, our method improves the success rate on all four tasks. The largest improvement occurs on the make watermelon juice task, where the success rate rises from 56.7\% to 83.3\%.

The results demonstrate that the proposed method can be deployed to complex real-world manipulation tasks involving multiple sequential subtasks and coordinated object interactions. In addition, the ablation variants $\pi_{0.5}$-Geometry and $\pi_{0.5}$-Global+Local Query achieve consistent gains over the $\pi_{0.5}$-SFT baseline, while our full method attains the highest success rate on all four tasks.

\section{CONCLUSIONS}

This paper presented a training-phase prior injection framework for fine-grained bimanual manipulation that combines physics-grounded flow-matching supervision via differentiable forward kinematics with a decoupled global--local query mechanism distilled from frozen visual teachers, while removing all auxiliary supervision modules at inference. Simulation and real-robot experiments on long-horizon bimanual tasks show that the two priors improve accuracy and robustness in a complementary, regime-dependent manner. They are super-additive in-distribution, whereas geometric supervision trades part of the randomization robustness for in-distribution precision. These findings also point to several limitations that suggest concrete directions for future work. The interaction between the two objectives is currently governed by static loss weights, and adaptively balancing them across deployment regimes may recover part of the traded-off robustness. The local priors depend on the quality of the offline detector, motivating more robust region-prior estimation. Finally, our evaluation is limited to a single simulation benchmark and one real dual-arm platform, which we plan to extend to broader embodiments and task suites.


\end{document}